\documentclass[10pt,journal]{IEEEtran}
\usepackage{amsmath,amsfonts}
\usepackage{algorithmic}
\usepackage{algorithm}
\usepackage{array}
\usepackage[caption=false,font=normalsize,labelfont=sf,textfont=sf]{subfig}
\usepackage{textcomp}
\usepackage{stfloats}
\usepackage{url}
\usepackage{verbatim}
\usepackage{graphicx}
\usepackage{cite}
\usepackage{amsmath}
\usepackage{amsthm}
\usepackage{multirow}
\usepackage{booktabs}
\usepackage{color}
\usepackage{makecell}
\newcommand{\etal}{\textit{et al}.}

\begin{document}

\title{Exploring Vulnerabilities of No-Reference Image Quality Assessment Models: A Query-Based Black-Box Method}

\author{Chenxi Yang, 
        Yujia Liu, 
        Dingquan Li, 
        Tingting Jiang %
}

\markboth{}%
{Yang \MakeLowercase{\etal}: Exploring Vulnerabilities of No-Reference Image Quality Assessment Models: A Query-Based Black-Box Method}

\maketitle

\begin{abstract}
No-Reference Image Quality Assessment (NR-IQA) aims to predict image quality scores consistent with human perception without relying on pristine reference images, serving as a crucial component in various visual tasks. Ensuring the robustness of NR-IQA methods is vital for reliable comparisons of different image processing techniques and consistent user experiences in recommendations. The attack methods for NR-IQA provide a powerful instrument to test the robustness of NR-IQA. However, current attack methods of NR-IQA heavily rely on the gradient of the NR-IQA model, leading to limitations when the gradient information is unavailable. In this paper, we present a pioneering query-based black box attack against NR-IQA methods. We propose the concept of \emph{score boundary} and leverage an adaptive iterative approach with multiple score boundaries. Meanwhile, the initial attack directions are also designed to leverage the characteristics of the Human Visual System (HVS). Experiments show our method outperforms all compared state-of-the-art attack methods and is far ahead of previous black-box methods. The effective NR-IQA model DBCNN suffers a Spearman's rank-order correlation coefficient (SROCC) decline of $0.6381$ attacked by our method, revealing the vulnerability of NR-IQA models to black-box attacks. The proposed attack method also provides a potent tool for further exploration into NR-IQA robustness. 
\end{abstract} %


\section{Introduction}
\IEEEPARstart{I}{mage} Quality Assessment (IQA) aims to predict image quality scores consistent with human perception, which can be categorized as Full-Reference (FR), Reduced-Reference (RR), and No-Reference (NR) according to the access to the pristine reference images. Among them, NR-IQA has witnessed substantial development recently and has emerged as a suitable method for real-world scenarios~\cite{su2020blindly,zhang2020blind} because it does not rely on reference images. NR-IQA models also serve as a crucial component in various visual tasks, such as evaluating image processing algorithms~\cite{gu2020image} and optimizing image recommendation systems~\cite{deng2014image}. The robustness of NR-IQA methods is vital for providing a stable and dependable basis for comparing different image processing techniques and ensuring consistent user experiences.

\begin{figure}[!t]
  \centering
  \includegraphics[width=.9\linewidth]{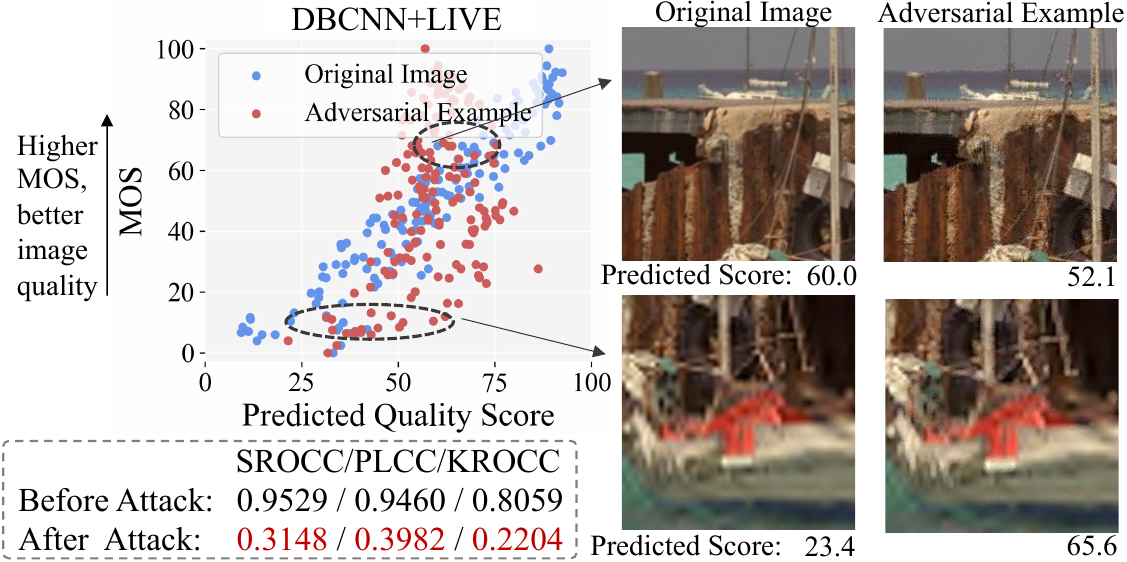}
  \caption{An NR-IQA model (DBCNN) is attacked in a black-box scenario. The top left shows the predicted quality scores by DBCNN on the LIVE dataset for original images and adversarial examples generated by our attack method. The bottom left shows the SROCC/PLCC/KROCC before and after the attack. The right figures show two sample images before and after the attack as well as the predicted scores.}\label{fig:method1}
\end{figure}

To scrutinize the robustness of NR-IQA models, recent research has conducted preliminary investigations, shedding light on the vulnerability of IQA models to various attacks~\cite{zhang2022perceptual,shumitskaya2022universal,korhonen2022adversarial}. These attack methods are designed to generate adversarial examples by causing significant deviations in the predicted quality scores from those of the original samples in two scenarios. In a \emph{white-box scenario} that the entire NR-IQA model under attack is available, generating adversarial examples using gradient-based optimization with the model's gradients is straightforward~\cite{zhang2022perceptual,shumitskaya2022universal,sang2023generation}. However, this white-box scenario becomes unrealistic when the model parameters are unknown to the attacker. In a \emph{black-box scenario}, attackers possess limited knowledge of the NR-IQA model, often confined to only its output. Korhonen~\etal~utilized a transfer-based method employing a substitute model to generate adversarial examples, which are then transferred to attack unknown target models~\cite{korhonen2022adversarial}. However, the performance of transfer-based black-box methods is limited, highly depending on the choice of substitute models and constraints~\cite{zhang2022perceptual}. 

This situation raises a challenging and intricate question: How can we attack IQA models in the black-box scenario, and without using substitute models? A potential resolution is to leverage query-based black-box attack methods, which are extensively explored in classification tasks~\cite{guo2019simple,li2022decision}. These approaches aim to design attack direction with stochasticity and prior information to generate adversarial examples that cross the classification boundary (\textit{i.e.} the prediction of the attacked model changed).

However, unlike widely studied black-box attacks for classification problems, attacking NR-IQA presents distinct challenges. Firstly, quantifying the success of attacks on regression-based IQA problems is not straightforward. Different from classification, which naturally defines a classification boundary for determining attack success, regression-based IQA lacks a direct measure of attack ``success'' due to its continuous output. Secondly, identifying the attack direction becomes particularly challenging when the gradient of an IQA model is unavailable. Unlike classification tasks, where a small perturbation like Gaussian noise may easily lead to successful attacks, the IQA problem demands a more deliberate design of the attack direction to generate substantial changes in the predicted quality scores. In our preliminary experiment where we attacked images for the classification and NR-IQA task with the Gaussian noise, the misclassification rate achieved $92.6\%$ but the quality score only changed by $2.09$ on average, with the predicted image quality score within the range of $[0,100]$. The result shows that the efficiency of this stochastic attack direction dropped dramatically in the context of NR-IQA. This disparity emphasizes the need for a more thoughtful and delicate design of attack directions in the context of NR-IQA. Thirdly, NR-IQA tasks are more sensitive to image quality variation than classification tasks. So attacking NR-IQA models has a more strict constraint for the perceptual similarity between the adversarial example and its original image, which implies that the perturbation is expected to be imperceptible for humans but could cause misjudgments by NR-IQA models. These intricate challenges underscore the significance of developing tailored black-box attack strategies for NR-IQA methods.

We address these three challenges in this paper. Firstly, the concept of \emph{score boundary} is introduced to quantify the success of individual attacks and systematically intensify attacks by setting multiple score boundaries, which enables a more measurable assessment of attack effectiveness. Secondly, leveraging the sensitivity of deep neural networks (DNNs) to texture information~\cite{ding2022image} and sparse noise~\cite{modas2019sparsefool}, we extract the texture and sparse noise from natural images and use them to design the attack direction. We constrain the attack region to the edges and salient areas of an image to enhance the efficacy of the attack. Thirdly, to ensure perturbation invisibility, we generate adversarial examples with the help of Just Noticeable Difference (JND)~\cite{liu2010just}. JND accounts for the maximum sensory distortion that the Human Visual System (HVS) does not perceive, and it provides a threshold for perturbation for each pixel in an image. When the perturbation of each pixel satisfies the constraint of the JND threshold, the perturbation of the whole image can be considered invisible to human eyes.
To optimize the final attack, we employ the SurFree framework~\cite{maho2021surfree}. This framework capitalizes the geometric properties of score boundaries and provides an effective query-based attack. %

The efficacy of our attack method is evaluated on four NR-IQA methods across two datasets.
The assessment employs three correlation metrics and Mean Absolute Error (MAE) to quantify the performance of the attack.  Additionally, two perceptual similarity metrics SSIM~\cite{wang2014image} and LPIPS~\cite{zhang2018unreasonable} are employed to measure the visibility of perturbations. We compare our approach to three transfer-based attack methods. The results demonstrate that while maintaining comparable invisibility of the perturbations, our method achieves superior attack effects. One intuitive case of our attack performance is shown in Fig.~\ref{fig:method1}.

Our contributions are as follows:
\begin{itemize}
    \item A novel query-based black box attack method against NR-IQA methods is proposed, featuring adaptive iterative attacks with initial attack direction guidance. To the best of our knowledge, this is the first work to design the query-based black-box attack for NR-IQA.
    \item We propose the concept of \emph{score boundary} for NR-IQA attacks and develop adaptive iterative score boundaries to adjust the attack intensity of different images. With prior knowledge of NR-IQA, we design initial attack directions based on the edge and salient areas of the attacked image. Besides, the constraint of JND is introduced, effectively reducing the visibility of the perturbation. 
    \item Extensive experiments show our attack achieves the best black-box performance on different NR-IQA methods, which reveals the vulnerability of NR-IQA under black-box attacks. Our exploration of black-box attacks on NR-IQA provides a convenient tool for further research of NR-IQA robustness.
\end{itemize}

\section{Related Work}
\subsection{Adversarial Attack in Classification Tasks}
Adversarial attack is an important problem considering the security and reliability of models. It has been studied extensively in classification, whose goal is to generate adversarial examples misclassified by the model, under the constraint of small perturbations around original images. It can be categorized into white-box attacks and black-box attacks. In white-box scenarios, attackers have access to all details of the target models, including their structures, parameters, and other relevant 
information~\cite{szegedy2014intriguing,baluja2018learning}. Most white-box attacks generate adversarial examples by solving a constrained optimization problem, where the constraint ensures the similarity between the original images and the generated examples. Commonly used conditions for this constraint include $\ell_\infty$ norm~\cite{2015_ICLR_Goodfellow_FGSM}, $\ell_2$ norm~\cite{2016_CVPR_Moosavi_DeepFool} and others~\cite{modas2019sparsefool,2020_CVPR_Zhao_ColorSpaceAttack}.

While in black-box scenarios, attackers possess little knowledge about the target model, often limited to just its output~\cite{chen2017zoo,papernot2016transferability}. In practical applications, black-box attacks are more common and challenging~\cite{xie2019improving}. 
There are two primary approaches for designing black-box attacks: transfer-based methods and query-based methods. Transfer-based methods first leverage known substitution models to generate adversarial examples, which are then transferred to attack unknown target models. %
Papernot~\etal~\cite{papernot2017practical} train a model to substitute for the target model, use the substitute to craft adversarial examples, and then transfer them to target models. 
On the other hand, query-based methods directly approximate the gradient by querying the target model and obtaining its output, allowing them to design adversarial perturbations based on the gradient. These methods do not require training a substitute model, focusing instead on direct interactions with the target model. For instance, Guo~\etal~\cite{guo2019simple} propose a strategy where adversarial examples are generated by iteratively adding or subtracting vectors from a predefined orthonormal basis, although this method requires a significant number of queries to ensure attack success. 
To address the inefficiency of high query demands, Thibault~\etal~\cite{maho2021surfree} introduce a method that capitalizes on the geometric properties of classifier decision boundaries to reduce the number of required queries. Meanwhile, using frequency mixup techniques, Li~\etal~\cite{li2022decision} effectively generate adversarial examples with limited queries.

\subsection{Image Quality Assessment}
IQA plays an important role in evaluating the perceptual quality of images, aiming to align closely with human visual judgment, commonly quantified as the Mean Opinion Score (MOS). IQA models strive to predict image quality in a cost-efficient manner compared to extensive human rating processes. 
Among them, FR-IQA measures the perceptual difference between the distorted image and its undistorted version, while NR-IQA predicts the image quality of a distorted image with no reference image.
Some IQA methods consider the signal-level information like luminance and edge in the spatial domain~\cite{wang2014image,zhang2011fsim} and natural scene statistics in the frequency domain~\cite{sheikh2006image,moorthy2011blind}. FSIM index~\cite{zhang2011fsim} is an exemplary method that quantifies image similarity, taking into account factors like chrominance and phase congruency between a pristine reference and a distorted image.
Furthermore, the image semantic information is also considered in the IQA task~\cite{li2019which,su2020blindly,song2023active,zhou2024deep}. The SFA method~\cite{li2019which} designs statistics derived from features extracted via neural networks trained on classification tasks. HyperIQA~\cite{su2020blindly} utilizes similar features to predict parameters of a quality prediction network. DDNet~\cite{zhou2024deep} used a dynamic filtering module to extract content-adaptive features. 
An additional dimension in IQA research considers the JND. It models the minimum visibility threshold of the HVS, as a critical component in several IQA methods~\cite{ferzli2009no,seo2021novel}. Recent years have also seen the exploration of unsupervised methods~\cite{madhusudana2022image,zhou2023blind}, multi-modality method~\cite{wang2023exploring}, and other innovative techniques, addressing the application scenarios appearing in recent years. 

Distortions in images are typically categorized into synthetic and authentic distortion. The former is artificially created and the latter occurs naturally during the image production process. Authentic distortion has a broader variety and complexity than synthetic distortion.

\subsection{Adversarial Challenges in Quality Assessment Tasks}
For attacking IQA, a general goal is to generate the adversarial example within a small perturbation around the original image while the image quality score changes a lot against the original image. In the white-box scenario, Zhang~\etal~\cite{zhang2022perceptual} employ a gradient-based optimization strategy, incorporating the Lagrangian method with a Full-Reference IQA (FR-IQA) as a perceptual constraint, to generate adversarial examples; Shumitskaya~\etal~\cite{shumitskaya2022universal} propose a universal adversarial perturbation to train a single adversarial perturbation applicable across an entire dataset. They further propose four different attack methods with universal adversarial perturbation to verify the adversarial robustness of IQA models in~\cite{shumitskaya2024towards}. 

In the black-box scenario, Korhonen and You~\cite{korhonen2022adversarial} utilize a substitute model to generate adversarial examples and then attack target models. The efficacy of such attacks is significantly influenced by the choice of the substitute model and the dataset used for training.
For example, when Zhang~\etal~\cite{zhang2022perceptual} demonstrate a notable variance in attack performance against the CORNIA model~\cite{ye2012unsupervised} depending on the substitute model employed, illustrating a disparity of $0.1151$ in the Spearman's Rank-Order Correlation Coefficient (SROCC) when using different substitute models (UNIQUE~\cite{zhang2021uncertainty} vs. BRISQUE~\cite{mittal2012no}).

This scenario underlines the necessity for black-box approaches that operate independently of substitute models, aiming to mitigate the reliance on specific training datasets and models. In the domain of Video Quality Assessment (VQA), Zhang~\etal~\cite{zhang2023vulnerabilities} propose a patch-based random search technique coupled with a score-reversed boundary loss for executing query-based black-box attacks on videos. While its score-reversed boundary loss provides effective guidance to the attack, there is still room for improving the efficiency due to its patch-based random search.

\section{Methodology}
\begin{figure}[!t]
  \centering
  \includegraphics[width=\linewidth]{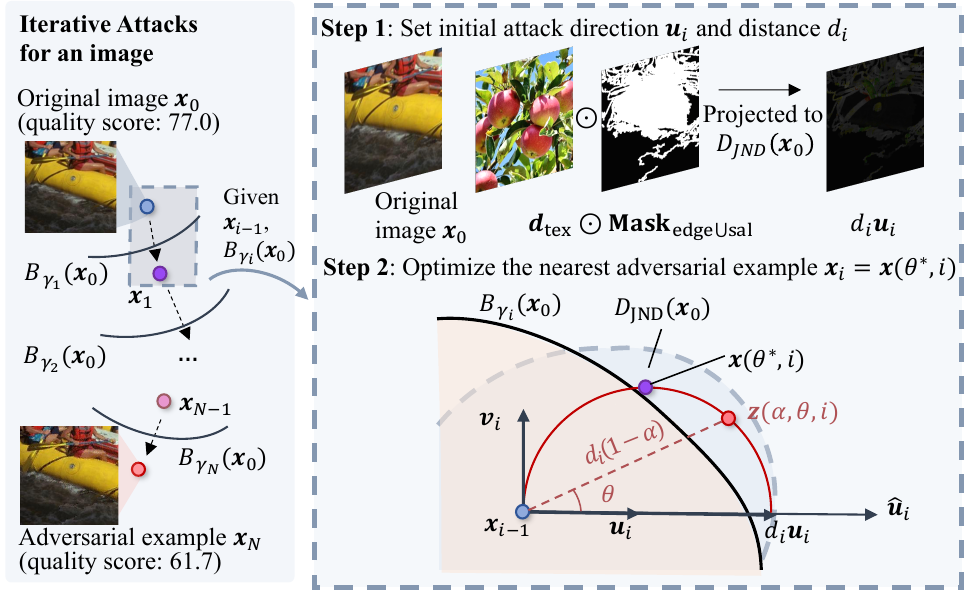}
  \caption{The framework of the proposed attack method.}\label{fig:main}
\end{figure}

In this section, we will introduce our method in a top-down order. We will first illuminate the Global-Local optimization objective of the entire attack process. Then the score boundary for a single-step attack and progressive score boundaries for multi-step attacks are defined. Finally, the optimization method with HVS prior for a single-step attack and adaptive score boundaries are described in detail. The framework of the attack is shown in Fig.~\ref{fig:main}.

\subsection{Global-Local Optimization Objective}\label{sec:GL}
To attack an NR-IQA model $f$, a primary goal is to make the predicted quality score $f(\boldsymbol{x}_0+\boldsymbol{\delta})$ of the attacked image deviate from the original score $f(\boldsymbol{x}_0)$ as much as possible, where $\boldsymbol{\delta}$ is the perturbation on the image $\boldsymbol{x}_0$. Meanwhile, the rank correlation of the predicted score with MOS is also important for NR-IQA, so we also hope to ``perceive'' rank correlation through adversarial samples. To disturb the rank correlation in an image set, we propose a Global-Local (GL) optimization objective for a more reasonable attack. For an image set $\mathcal{I}$, we split it into the higher quality part $\mathcal{I}_h$ and the lower quality part $\mathcal{I}_l$ according to the original image score $f(\boldsymbol{x}_0)$. Our goal involves inducing the model to misjudge a high-quality image with a lower quality, and vice versa. A larger $f(\boldsymbol{x}_0)$ corresponds to a higher quality of $\boldsymbol{x}_0$. The optimization objective is
\begin{equation}
\begin{gathered}
    \max_{\boldsymbol{\delta}}
    S(\boldsymbol{x}_0)*\left(f(\boldsymbol{x}_0+\boldsymbol{\delta})- f\left(\boldsymbol{x}_0\right)\right), \\
    \text { s.t. } \boldsymbol{x}_0+\boldsymbol{\delta}\in \mathcal{D}_{\text{JND}}(\boldsymbol{x}_0),
\label{eq2}
\end{gathered}
\end{equation}
where 
\begin{equation}
    \begin{aligned}
    S(\boldsymbol{x}_0)=\left\{
    \begin{array}{ll}
    1, &\boldsymbol{x}_0 \in \mathcal{I}_l \\
    -1, &\boldsymbol{x}_0 \in \mathcal{I}_h
    \end{array}.\right.
    \end{aligned}
    \label{eq3}
\end{equation}
And to restrain the visibility of perturbation, the adversarial sample $\boldsymbol{x}_0+\boldsymbol{\delta}$ is restrained to the JND neighborhood $\mathcal{D}_{\text{JND}}(\boldsymbol{x}_0)$ of $\boldsymbol{x}_0$.
The neighborhood $D_{\text{JND}}\left(\boldsymbol{x}_0\right)$ of an image $\boldsymbol{x}_0$ could be written as:
\begin{equation}
\resizebox{.88\linewidth}{!}{$
\begin{aligned}
    D_{\text{JND}}\left(\boldsymbol{x}_0\right)=\left\{\boldsymbol{x}\big\vert\left|\boldsymbol{x}(l,j,k)-\boldsymbol{x}_0(l,j,k)\right|<m(l,j,k),\right. \\ %
    \left. 0\leq l<H,0\leq j<W,0\leq k<C\right\},
\end{aligned}$}
 \label{eq7}
\end{equation}
where $m(l,j,k)$ is the minimum visibility threshold at pixel $\boldsymbol{x}_0(l,j,k)$ located on $(l,j,k)$ on image $\boldsymbol{x}_0$ predicted by a JND model. The height, weight, and channel of $\boldsymbol{x}_0$ are $H,W$, and $C$ respectively. The JND model is to estimate the pixel-wise threshold for an image, the perturbed image cannot be visually distinguished from the original image $\boldsymbol{x}_0$ if the perturbation is under the threshold~\cite{liu2010just}.

\subsection{Iterative Score Boundaries for Optimization}\label{sec:iter}
To qualify the variation of the predicted quality score during attacking, we propose the concept of \emph{score boundary}. For example, for an image $\boldsymbol{x}_0\in \mathcal{I}_l$, and maximum and minimum MOS value $\text{MOS}_\text{max},\text{MOS}_\text{min}$ in the dataset, we set the score boundary $B_{\gamma}^l(\boldsymbol{x}_0)$ as
\begin{equation}
  B_{\gamma}^l(\boldsymbol{x}_0)=\{\boldsymbol{x}|f(\boldsymbol{x})=f(\boldsymbol{x}_0)+\gamma(\text{MOS}_\text{max}-f(\boldsymbol{x}_0))\}.  
\end{equation}
 This boundary includes samples with a higher quality score than $\boldsymbol{x}_0$. Then the attack is $\gamma$-success if $f(\boldsymbol{x}_0+\boldsymbol{\delta})>f(\boldsymbol{x}_0)+\gamma(\text{MOS}_\text{max}-f(\boldsymbol{x}_0))$. And $\gamma$ is a scalar to adjust the distance from $\boldsymbol{x}_0$ to $B_{\gamma}^l(\boldsymbol{x}_0)$, which corresponds to the attack intensity. While for $\boldsymbol{x}_0\in \mathcal{I}_h$, $B_{\gamma}^h(\boldsymbol{x}_0)=\{\boldsymbol{x}|f(\boldsymbol{x})=f(\boldsymbol{x}_0)+\gamma(\text{MOS}_\text{min}-f(\boldsymbol{x}_0))\}$. So we define the adversarial example $\boldsymbol{x}_0+\boldsymbol{\delta}$ is $\boldsymbol{\gamma}$\textbf{-success} if: 
\begin{equation}\label{eq_succ}
  \resizebox{.91\linewidth}{!}{$
    f(\boldsymbol{x}_0+\boldsymbol{\delta})\!\left\{
    \begin{array}{ll}
    \!>f(\boldsymbol{x}_0)+\gamma(\text{MOS}_\text{max}-f(\boldsymbol{x}_0)), \boldsymbol{x}_0 \in \mathcal{I}_l \\
    \!<f(\boldsymbol{x}_0)+\gamma(\text{MOS}_\text{min}-f(\boldsymbol{x}_0)), \boldsymbol{x}_0 \in \mathcal{I}_h
    \end{array}.\right.$
  }
\end{equation}
With the criterion in Eq.~\eqref{eq_succ}, the success of a single-step attack with intensity $\gamma$ could be obtained. 

Further, to determine the maximum attack intensity of an image, multiple score boundaries are applied. For $\boldsymbol{x}_0\in \mathcal{I}_l$ and initial $\gamma_0,\gamma_{-1}$, a series of score boundaries $B_{\gamma_1}^l(\boldsymbol{x}_0),..., B_{\gamma_N}^l(\boldsymbol{x}_0)$ are set with $\gamma_i=\gamma_{i-1}+(\gamma_{i-1}-\gamma_{i-2}),i=1,...,N$. With the multi-step attacks, a series of adversarial images $\boldsymbol{x}_1,...,\boldsymbol{x}_N$ could be generated, which satisfy the property that $\boldsymbol{x}_i$ is $\gamma_i$-success ($i=1,...N$). And $\boldsymbol{x}_N$ is used as the final adversarial example for $\boldsymbol{x}_0$. The algorithm for iterative attacks is shown in Algorithm~\ref{alg:algorithm_all}.
The iterative boundaries guarantee the attack intensity of each iteration is moderate, while multiple boundaries ensure the considerable attack intensity of the whole attack. Further adaptive optimization for iterative score boundaries will be shown in Sec.~\ref{sec:ada}.

\begin{algorithm}[!t]
\caption{Algorithm for Iterative Attacks}
\label{alg:algorithm_all}
\textbf{Input}: Original image $\boldsymbol{x}_{0}$, maximum number of score boundaries $N$, initial $\gamma_0=1/100$, $\gamma_{-1}=0$\\
\textbf{Output}: Adversarial point $\boldsymbol{x}_{N}$
\begin{algorithmic}[1] %
\FOR{$i\leftarrow1,...,N$}
\STATE $\gamma_i\leftarrow\gamma_{i-1}+(\gamma_{i-1}-\gamma_{i-2})$ %
\STATE $\boldsymbol{x}_i,\gamma_i\leftarrow SingleAttack(\boldsymbol{x}_{i-1},\gamma_i,...)$ // Algorithm~\ref{alg:algorithm}
\ENDFOR
\RETURN $\boldsymbol{x}_N$
\end{algorithmic}
\end{algorithm}

\begin{algorithm}[tb]
\caption{$SingleAttack$ (Algorithm for A Single-Step Attack)}
\label{alg:algorithm}
\textbf{Input}: Start point $\boldsymbol{x}_{i-1}$, original image $\boldsymbol{x}_0$, JND neighbourhood $D_{\text{JND}}\left(\boldsymbol{x}_0\right)$, score boundary $B_{\gamma_i}(\boldsymbol{x}_0)$, image $\boldsymbol{d}_{\text{tex}}$, maximum search times for a single-step attack $T_\text{max}$ \\
\textbf{Output}: Adversarial point $\boldsymbol{x}_{i}$
\begin{algorithmic}[1] %
\STATE Search times $T\leftarrow0$
\STATE Generate $\textbf{Mask}_{\text{edge}\cup \text{sal}}(\boldsymbol{x}_0)$ \\
\STATE Set initial attack direction\\
$\boldsymbol{\hat{u}_i}\leftarrow\tau\cdot\boldsymbol{d}_{\text{tex}}\odot\textbf{Mask}_{\text{edge}\cup \text{sal}}(\boldsymbol{x}_0),\tau\sim U(-0.1,0.1)$.
\STATE $d_i \leftarrow ||\text{Proj}_{D_{\text{JND}}(\boldsymbol{x}_0)}(\boldsymbol{\hat{u}_i})||, \boldsymbol{u}_i\leftarrow\text{Proj}_{D_{\text{JND}}(\boldsymbol{x}_0)}(\boldsymbol{\hat{u}_i})/d_i$
\IF {$\boldsymbol{x}_{i-1}+d_i\boldsymbol{u}_i$ is not $\gamma_i$-success \textbf{or} $d_i<1$}
\STATE $T\gets T+1$
\IF {$T>T_\text{max}$}
\STATE // To decrease $\gamma_i$
\STATE $\gamma_i \leftarrow \gamma_{i}-(\gamma_i-\gamma_{i-1})/2$, go to line {\textcolor{red}{1}} %
\ELSE 
\STATE go to line {\textcolor{red}{2}}
\ENDIF
\ENDIF
\IF {$\boldsymbol{x}_{i-1}+d_i\boldsymbol{u}_i$ is $(\gamma_i+2(\gamma_i-\gamma_{i-1}))$-success}
\STATE // To increase $\gamma_i$
\STATE $\gamma_i\leftarrow \gamma_i+(\gamma_i-\gamma_{i-1})$
\ENDIF
\STATE Set another stochastic attack direction $\boldsymbol{{v}}_i$
\STATE $\boldsymbol{x}(\theta,i) \leftarrow d_i \cos \theta\left(\boldsymbol{u}_i\cos \theta +\boldsymbol{v}_i\sin \theta \right)+\boldsymbol{x}_{i-1}$
\STATE $\theta^*\leftarrow\underset{\theta, \boldsymbol{x}(\theta,i) \text{ is }\gamma_i\text{-success}}{\operatorname{argmin}}\left\|\boldsymbol{x}(\theta,i)-\boldsymbol{x}_{i-1}\right\|$ %
\STATE $\boldsymbol{x}_i \leftarrow \text{Proj}_{D_{\text{JND}(\boldsymbol{x}_0)}}{(\boldsymbol{x}(\theta^*,i))}$
\RETURN $\boldsymbol{x}_i,\gamma_i$
\end{algorithmic}
\end{algorithm}

\subsection{Optimization Method for A Single-Step Attack}
\label{sec:single}
With the target decomposition in Sec.~\ref{sec:iter}, the attack objective of the $i$th-step attack of $\boldsymbol{x}_0$ could be set with:

Find $\boldsymbol{x}_i\in \mathcal{D}_{\text{JND}}(\boldsymbol{x}_0)$, subject to $\boldsymbol{x}_i$ is $\gamma_i$-success.

To solve this problem, we leverage a query-based black-box method~\cite{maho2021surfree} for classification attack, which reaches low query amounts in attacking classification tasks by utilizing geometrical properties of the classiﬁer decision boundaries. In our attack on NR-IQA, the same analysis could be used. With a start point $\boldsymbol{x}_{i-1}$, a preset unit attack direction $\boldsymbol{u}_i$ and a distance $d_i$ which satisfies $\boldsymbol{x}_{i-1}+d_i\boldsymbol{u}_i$ is $\gamma_i$-success, and a stochastic unit direction $\boldsymbol{v}_i$ orthogonal to $\boldsymbol{u}_i$, the polar coordinate of a point $\boldsymbol{z}$ near $\boldsymbol{x}_{i-1}$ could be represented as
\begin{equation}
    \boldsymbol{z}(\alpha,\theta,i)=d_i(1-\alpha)(\boldsymbol{u}_i\cos\theta+\boldsymbol{v}_i\sin\theta)+\boldsymbol{x}_{i-1},
\end{equation}
where $\alpha\in[0,1], \theta\in[-\pi,\pi]$. Given $\alpha$, the trajectory of $\boldsymbol{z}(\alpha,\theta, i)$ is an arc, which is shown as the red arc in the lower part of Fig.~\ref{fig:main} for $\theta\in[0,\pi/2]$. The goal is to choose $(\alpha,\theta)$ to raise the probability of $\boldsymbol{z}(\alpha,\theta, i)$ being adversarial. With the theoretical analysis in~\cite{maho2021surfree}, when $\alpha=1-\cos\theta$, probability of $\boldsymbol{z}(\alpha,\theta,i)$ being adversarial reaches maximum.
So we mark $\boldsymbol{z}(1-\cos\theta,\theta,i)$ as the candidate point $\boldsymbol{x}(\theta,i)$:
\begin{equation}
    \boldsymbol{x}(\theta,i) =  d_i\cos \theta\left(\boldsymbol{u}_i\cos \theta +\boldsymbol{v}_i\sin \theta \right)+\boldsymbol{x}_{i-1}.
    \label{eq4}
\end{equation} %

The adversarial example $\boldsymbol{x}_{i}:=\boldsymbol{x}(\theta,i)$ can be solved by
\begin{equation}
\begin{aligned}
    &\underset{\boldsymbol{x}(\theta,i)}{\operatorname{min}}\left\|\boldsymbol{x}(\theta,i)-\boldsymbol{x}_{i-1}\right\|,\\
    \text{ s.t. } \boldsymbol{x}(\theta,i)&\in D_{\text{JND}}\left(\boldsymbol{x}_0\right), \boldsymbol{x}(\theta,i)\text{ is }\gamma_i\text{-success}.
\end{aligned}
    \label{eq6}
\end{equation}
There are three questions in attacking NR-IQA: 1) The reasonable preset direction $\boldsymbol{u}_i$ should be deliberately designed to guarantee an efficient attack. 2) The preset direction $\boldsymbol{v}_i$ should be designed to guarantee the orthogonality with $\boldsymbol{u}_i$. 
3) The generated adversarial example should satisfy the constraint of $D_{\text{JND}}$, which is difficult in solving Eq.~\eqref{eq6}.

For the design of attack direction $\boldsymbol{u}_i$, in attacking classification tasks, a common approach is to employ stochastic perturbations, such as Gaussian noise, as the attack direction $\boldsymbol{u}_i$ in Eq.~\eqref{eq4}. However, this strategy, while effective for classification tasks, often proves inadequate when targeting NR-IQA models.
For instance, when applying the attack perturbation $\boldsymbol{\delta}$ with preset values $\boldsymbol{\delta}\sim0.15\cdot\mathcal{N}(0,1)$ to the starting point $\boldsymbol{x}_0$ (normalized to the range $[0,1]$) of the classification task, we observe a high success rate of $92.6\%$ for inducing misclassification of $\boldsymbol{x}_0+\boldsymbol{\delta}$ in $500$ random trials. However, when utilizing the same attack direction to target the NR-IQA model HyperIQA, the resulting average change between $f(\boldsymbol{x}_0)$ and $f(\boldsymbol{x}_0+\boldsymbol{\delta})$ is merely $2.09$ (within the range of predicted image scores $[0,100]$). The efficiency of this stochastic attack direction dropped dramatically in the context of NR-IQA.
In our pursuit of a more effective attack perturbation, we introduce a method that disrupts image regions that are sensitive to NR-IQA models while ensuring the perturbation remains invisible to the human eye.%

\subsubsection{Designing the preset direction $\boldsymbol{u}_i$} 
We generate $\boldsymbol{u}_i$ with three steps. Firstly, we design an initial perturbation $\boldsymbol{d}_{\text{tex}}$, which contains image texture and sparse noise. Secondly, the perturbation $\boldsymbol{\hat{u}}_i$ is obtained by confining $\boldsymbol{d}_{\text{tex}}$ to special image regions. Finally, the attack direction $\boldsymbol{u}_i$ is obtained by applying the projection and normalization operation on $\boldsymbol{\hat{u}}_i$. 

In the first step, regarding the initial perturbation $\boldsymbol{d}_\text{tex}$, we are inspired by existing work exploring
the sensitivity of DNNs to both image texture~\cite{ding2022image} and sparse noise~\cite{modas2019sparsefool,dong2020greedyfool}, and leverage the texture information and sparse noise extracted from the high-quality natural images $\boldsymbol{I}_\text{nat}$. The extracted information is denoted as high-frequency information $\boldsymbol{I}_\text{hreq}=g(\boldsymbol{I}_\text{nat})$, with an extraction function $g(\cdot)$. Meanwhile, $\boldsymbol{d}_\text{tex}$ is crafted to match the dimensions of the attacked image $\boldsymbol{x}_{i-1}$. More options for $\boldsymbol{d}_\text{tex}$ are explored in Sec.~\ref{sec_dtex_option}.

In the second step, for designing $\boldsymbol{\hat{u}}_i$, our idea is to add disruption to the sensitive image regions for NR-IQA models, while the disruption is not visible to the human eye. Noting that the edge region and salient region are often critical to the judgment of IQA models~\cite{zhang2011fsim,seo2021novel}, we introduce a mask $\textbf{Mask}_{\text{edge}\cup \text{sal}}(\boldsymbol{x}_0)$ to confine attacks to these specific regions, whose role is to preserve perturbations in the edge and salient regions of $\boldsymbol{x}_0$, and remove perturbations in other regions.
The designed perturbation could be formulated as 
$\boldsymbol{\hat{u}}_i=\tau\cdot\boldsymbol{d}_{\text{tex}}\odot\textbf{Mask}_{\text{edge}\cup \text{sal}}(\boldsymbol{x}_0)$,
where $\odot$ is the Hadamard product, $\tau$ is a stochastic scalar drawn from a uniform distribution $U(-0.1,0.1)$. The $\tau$ introduces different intensities in searching for $\boldsymbol{\hat{u}}_i$, enhancing the versatility and adaptability of the proposed method.

In the third step, to obtain the initial attack direction $\boldsymbol{u}_i$, $\boldsymbol{\hat{u}}_i$ is firstly modified by a projection to $D_{\text{JND}}$. The projection operation is defined as:
\begin{equation}
    \text{Proj}_{D_{\text{JND}}(\boldsymbol{x}_0)}(\boldsymbol{\hat{u}}_i):=\underset{\boldsymbol{\tilde{u}}, \boldsymbol{x}_{i-1}+\boldsymbol{\tilde{u}}\in D_{\text{JND}}(\boldsymbol{x}_0)}{\operatorname{argmin}}{\left\|\boldsymbol{\tilde{u}}-\boldsymbol{\hat{u}}_i\right\|}.
    \label{eq:proju}
\end{equation}
Then the resulting projected vector is then normalized to obtain $\boldsymbol{u}_i$:
\begin{equation}
   \begin{aligned}
   d_i=&||\text{Proj}_{D_{\text{JND}}(\boldsymbol{x}_0)}(\boldsymbol{\hat{u}}_i)||,\\
   \boldsymbol{u}_i=&\frac{\text{Proj}_{D_{\text{JND}}(\boldsymbol{x}_0)}(\boldsymbol{\hat{u}}_i)}{d_i}.
   \end{aligned}
   \label{eq:diui}
\end{equation}
To ensure $\boldsymbol{u}_i \in D_{\text{JND}}$, any $\boldsymbol{\hat{u}}_i$ with $d_i<1$ is discarded and a new $\boldsymbol{\hat{u}}_i$ is regenerated.

\subsubsection{Designing the preset direction $\boldsymbol{v}_i$} 
For $\boldsymbol{v}_i$, we follow the practice in~\cite{maho2021surfree} and generate $\boldsymbol{{v}}_i$ with the stochastic sample on the low-frequency subband of the original image.
Firstly the image is transformed to the frequency domain with the full Discrete Cosine Transform (DCT) as in~\cite{li2020qeba}. Then a fraction $\rho$ of the transform coefficients is selected in the low-frequency subband. These selected transform coefﬁcients are reassigned values uniformly distributed over $\{-1,0,1\}$, while the remaining coefficients are set to 0. The inverse DCT transform yields the direction $\boldsymbol{{v}}_i$. Then, to guarantee the orthogonality between $\boldsymbol{{u}}_i$ and $\boldsymbol{{v}}_i$, the Gram-Schmidt process~\cite{Cheney2009linear} is employed.

\subsubsection{Generation of adversarial examples with $D_\text{JND}$}
With $\boldsymbol{u}_i$ and $\boldsymbol{v}_i$, Eq.~\eqref{eq6} could be solved with a binary search of $\theta$ to 
\begin{equation}
    \boldsymbol{{x}}_i=\text{Proj}_{D_{\text{JND}}(\boldsymbol{x}_0)}(\boldsymbol{x}(\theta^*,i)):=\underset{\boldsymbol{\tilde{x}}, \boldsymbol{\tilde{x}}\in D_{\text{JND}}(\boldsymbol{x}_0)}{\operatorname{argmin}}{\left\|\boldsymbol{\tilde{x}}-\boldsymbol{x}(\theta^*,i)\right\|}.
    \label{eq:projxi}
\end{equation}
The algorithm for a single-step attack is shown in Algorithm~\ref{alg:algorithm}.

\subsection{Adaptive Optimization for Score Boundaries}\label{sec:ada}
To fine-tune attack intensity for different images, we leverage an adaptive optimization for iterative score boundaries to set adjustable $\{\gamma_i\}_{i=0}^N$, which means the score boundaries are adaptive for each image and each iteration. When the boundary is too difficult to cross, a closer boundary with a smaller $\gamma$ is set. When the boundary is too easy to cross, a more distant boundary with a larger $\gamma$ is set. The benefit of adaptive boundaries is to guarantee a stronger attack, by adjusting the score boundary dynamically.

For two neighboring score boundaries $\gamma_{i-1},\gamma_{i}$ of an image, there are \emph{Decreasing} and \emph{Increasing strategies}: a) \emph{Decreasing strategy}: when maximum search times for initial attack direction $\boldsymbol{u}_i$ is achieved in a single-step attack, we decrease $\gamma_i$ to $\gamma_i-(\gamma_i-\gamma_{i-1})/2$.  b) \emph{Increasing strategy}: when initial attack direction $\boldsymbol{u}_i$ and distance ${d_i}$ satisfy that $\boldsymbol{x}_{i-1}+d_i\boldsymbol{u}_i$ is $(\gamma_i+2(\gamma_i-\gamma_{i-1}))$-success, increase $\gamma_i$ to $(\gamma_i+(\gamma_i-\gamma_{i-1}))$. These strategies are outlined in lines 9 and 16 of Algorithm~\ref{alg:algorithm}.
When the $\gamma_i$ is decreased and the difference $\gamma_i-\gamma_{i-1}<1/400$, the attack will be early stopped. This indicates that the attack intensity is nearing saturation in recent iterations. %

\section{Experiments}
In this section, we first present the setting of attacks, including attacked NR-IQA methods, and the experimental results compared with other methods. Then the effect of different parts of our attack is explored. Additionally, the visualization of adversarial examples is presented. Finally, the comparisons with other attack methods are shown.

\subsection{Experimental Setups}

\subsubsection{NR-IQA Models and Datasets}
We choose four NR-IQA models DBCNN~\cite{zhang2020blind}, HyperIQA~\cite{su2020blindly}, SFA~\cite{li2019which}, and CONTRIQUE~\cite{madhusudana2022image}, which are based on the various quality features extracted by DNN and are all widely recognized in the NR-IQA field.
The LIVE dataset~\cite{sheikh2006statistical} with synthetic distortions and CLIVE dataset~\cite{ghadiyaram2016massive} with authentic distortions are chosen to train and attack NR-IQA models respectively. %
$80\%$ data of the dataset are split for training and the rest for testing and the attack. No image content overlaps between the training and the test set. NR-IQA models are retrained on LIVE and CLIVE with their public code. The predicted scores are normalized to $[0,100]$.
For the attack, we use a random cropping with $224\times224$ for each image. And cropped images are fixed for all experiments.

\subsubsection{Setting of Attacking Experiments}
We set the number of score boundaries $N=20$, with $\gamma_0=0.01$ for $B_{\gamma_0}(\boldsymbol{x}_0)$. Maximum search times $T_\text{max}$ is set to $200$, $\text{MOS}_\text{max}=100$ and $\text{MOS}_\text{min}=0$. The $\mathcal{I}_h$ and $\mathcal{I}_l$ are split by whether $f(\boldsymbol{x}_0)$ exceeds $50$.
The saliency maps of $\boldsymbol{x}_0$ is predicted with MBS~\cite{zhang2015minimum}, and edges of $\boldsymbol{x}_0$ are extracted by Canny operation~\cite{canny1986computational}. In $\textbf{Mask}_{\text{edge}\cup \text{sal}}(\boldsymbol{x}_0)$, the pixel with a positive value in the salient map or edge map of $\boldsymbol{x}_0$ is set to $1$ and other pixels are set to $0$. The JND model of Liu~\etal~\cite{liu2010just} is used, which predicts a single-channel JND map of an image $\boldsymbol{x}_0$. We subsequently apply this JND map on each color channel of the image, as the $D_{\text{JND}}(\boldsymbol{x}_0)$.
For the convenience of optimization, the norm in the optimization target of Eq.~\eqref{eq:proju} and~\eqref{eq:projxi} is set to $L_2$ norm.
For $\boldsymbol{I}_\text{hfre}$, two high-quality images I60 and I71 are selected from the KADID-10k dataset~\cite{lin2019kadid} as $\boldsymbol{I}_\text{nat}$. 
The high-frequency information $\boldsymbol{I}_\text{hfre}$ is obtained by:
\begin{equation}
    \boldsymbol{I}_\text{hfre}=g(\boldsymbol{I}_\text{nat})=\boldsymbol{I}_\text{nat} - g_\text{blur}(\boldsymbol{I}_\text{nat}),
\end{equation}
where $g_\text{blur}(\cdot)$ is a Gaussian blur operation with a $3\times3$ kernel. 
The select $\boldsymbol{I}_\text{nat}$ and their extracted $\boldsymbol{I}_\text{hfre}$ are shown in the first two images in the first row and second rows of Fig.~\ref{fig:kadid}. 
For each single-step attack, one of two $\boldsymbol{I}_\text{hfre}$ is randomly selected as the initial attack direction $\boldsymbol{d}_{\text{tex}}$. The $\rho$ for the generation of $\boldsymbol{v}_i$ is set to $0.5$ in our experiments. For the whole attack of an image, the maximum number of queries is limited to 8000. %

\subsubsection{Evaluation of Attack Performance} To evaluate the attack performance, we consider the effects of attacks on both individual images and a set of images. For a single image, the absolute error between the predicted score of the adversarial example and MOS is calculated, and it is presented for the whole test set as MAE. For a set of images, we analyze the correlation between the predicted quality scores and MOS in the test set, employing three correlation indices: SROCC, Pearson linear correlation coefficient (PLCC), and Kendall rank-order correlation coefficient (KROCC). 
SROCC measures the monotonicity of the relation between MOS and predicted quality scores. PLCC measures the linear correlation between MOS and predicted quality scores, which accounts for the prediction accuracy. And KROCC measures the rank correlation with pairwise comparison.
For $M$ images within an image set, with MOS values represented as $l_1,...,l_M$ and predicted scores as $f_1,...,f_M$, the correlation indices are calculated as follows:
\begin{align}
    \text{SROCC}=&1-\frac{6\sum_{i=1}^Md_i^2}{M(M^2-1)}, \\
    \text{PLCC}=&\frac{\sum_{i=1}^M(l_i-\mu_l)(f_i-\mu_f)}{\sqrt{\sum_{i=1}^M(l_i-\mu_l)^2(f_i-\mu_f)^2}}, \\
    \text{KROCC}=&\frac{2(F_\text{con}-F_\text{dis})}{M(M-1)},
\end{align}
where $d_i$ is the rank difference between the MOS and predicted score for $i${th} image. $\mu_l$ and $\mu_f$ denote the mean values of MOS and predicted scores, respectively.  And $F_\text{con}$ and $F_\text{dis}$ indicate the count of concordant and discordant pairs. in the image set.
Furthermore, we introduce an analysis of robustness, $R$, as initially proposed by Zhang~\etal~\cite{zhang2022perceptual}, to evaluate the variation in predicted quality scores before and after the attack:
\begin{equation}
R=\frac{1}{M} \sum_{i=1}^M \log \left(\frac{\max \left\{f_i-\beta_1, \beta_2-f_i\right\}}{\left|f_i-f_i^{\star}\right|}\right),
\end{equation}
where $f_i$ and $f_i^*$ are the predicted scores of the original and attacked version of $i${th} image be attacked. $M$ is the total number of attacked images. And $\beta_1$ and $\beta_2$ correspond to the minimum and maximum values of MOS in the image set. A smaller $R$ value corresponds to a stronger attack for the attacker. In our experiments, $\beta_1, \beta_2$ are 3.42, 92.43 for the LIVE dataset, and 3.50, 90.55 for the CLIVE dataset.
For the invisibility performance, we use SSIM~\cite{wang2014image} and LPIPS~\cite{zhang2018unreasonable} to calculate the perceptual similarity between original images and adversarial examples.

\begin{table*}[htbp]
\setlength\tabcolsep{3.2pt}
\centering
\caption{Black-Box attack performances on four NR-IQA models. The best and second attack performances are marked with \textbf{bold} and \underline{underline}.
} %
\resizebox{\textwidth}{!}{%
\begin{tabular}{clcccc|cc|cccc|cc}
\toprule[0.4mm]
 & \multicolumn{1}{c}{} & \multicolumn{6}{c|}{LIVE} & \multicolumn{6}{c}{CLIVE} \\ \cline{3-14}
& \multicolumn{1}{c}{} & \multicolumn{4}{c|}{Attack Performance} & \multicolumn{2}{c|}{Invisibility} & \multicolumn{4}{c|}{Attack Performance} & \multicolumn{2}{c}{Invisibility} \\ \cline{3-14}
\multirow{-3}{*}{\begin{tabular}[c]{@{}c@{}}Attacked \\ NR-IQA\end{tabular}} 
& \multicolumn{1}{c}{\multirow{-3}{*}{\begin{tabular}[c]{@{}c@{}}Attack \\ Method\end{tabular}}} & SROCC$\downarrow$  & PLCC$\downarrow$ & KROCC$\downarrow$ & MAE$\uparrow$ & SSIM$\uparrow$ & LPIPS$\downarrow$ & SROCC$\downarrow$  & PLCC$\downarrow$  & KROCC$\downarrow$  & MAE$\uparrow$ & SSIM$\uparrow$  & LPIPS$\downarrow$   \\ \hline
& Original  & 0.9529          & 0.9460           & 0.8058          & 9.79          & -            & -     &
0.8133 &	0.8467 &	0.6292 &	8.39 & - & - \\
& Korhonen & 0.8766          & 0.8671          & 0.6928          & \underline{19.43}         & 0.867                        & 0.186
&\underline{0.6799}	& \underline{0.6856} &	\underline{0.4986} &	\underline{14.73} & 0.865 &	0.113\\
& UAP  & \underline{0.8311}          & \underline{0.8145}        & \underline{0.6409}          & 17.38        & 0.792     & 0.141  
& 0.7026 &	0.7083 &	0.5196 &	13.10 & 0.650 &	0.159 \\
& Zhang  & -               & -               & -               & -              & -  & - 
& -   & -   & -    & -  & -  & -  \\
\multirow{-5}{*}{DBCNN}   & Ours                                                & \textbf{0.3985} & \textbf{0.4952} & \textbf{0.2802} & \textbf{20.93} & {0.867}               & {0.082} 
& \textbf{0.1148} &	\textbf{0.1943} &	\textbf{0.0854} &	\textbf{18.25} & 0.891 &	0.100 \\ \hline
& Original    & 0.9756          & 0.9746          & 0.8714          & 4.09          & -         & -               
&0.8543 &	0.8816 &	0.6762 &	8.13 & - & -\\
 & Korhonen%
& 0.9161          & 0.9007          & 0.7417          & \underline{12.33}          & 0.867   & 0.186 
&\underline{0.6652} &	\underline{0.6630} &	\underline{0.4861} &	\underline{14.38} & 0.865	& 0.113\\
& UAP %
& \underline{0.8685}        & \underline{0.8525}          & \underline{0.6959}            & 10.73          & 0.792    & 0.141  
& 0.7847	& 0.7750	& 0.5909 &	10.89 & 0.650 & 0.159\\
& Zhang %
& 0.9664          & 0.9557          & 0.8421          & 5.99           & 0.853    & 0.084 
& 0.8208 &	0.8428 &	0.6347 &	9.26 & 0.791 &	0.113 \\
\multirow{-5}{*}{HyperIQA}  & Ours & \textbf{0.8334} & \textbf{0.8293} & \textbf{0.6525} & \textbf{13.01} & {0.868}  & {0.065}   
& \textbf{0.4055} & \textbf{0.5240} & \textbf{0.2823} & \textbf{14.59} & 0.879 &	0.111 \\ \hline
& Original   & 0.8425          & 0.8379          & 0.6546          & 11.99          & -                            & -     
&0.8050 &	0.8322 &	0.6205 &	9.23 & - & -\\
& Korhonen & 0.7479          & 0.7606          & 0.5585          & \textbf{21.22} & 0.867  & 0.186 
& \underline{0.5967} &	\underline{0.6029} &	\underline{0.4249} &	\textbf{22.33} & 0.865	& 0.113\\
  & UAP   & \underline{0.7458} &	\underline{0.7435} &	\underline{0.5469}	& 13.58         & 0.792   & 0.141
  & 0.5999 &	0.6307 &	0.4257 &	14.92 & 0.650 &	0.159\\
& Zhang    & 0.8519          & 0.8404          & 0.6612         & 11.33          & 0.853     & 0.084 
&0.7534 &	0.7912	& 0.5677 &	10.79 & 0.791 &	0.113 \\
\multirow{-5}{*}{SFA} & Ours & \textbf{0.5953} &	\textbf{0.6578} &	\textbf{0.4335} &	\underline{15.47}          & {0.867}               & {0.079} 
& \textbf{0.3368} &	\textbf{0.3818} &	\textbf{0.2329} &	\underline{16.52} & 0.882 &	0.101 \\ \hline                                 
& Original & 0.8682	& 0.8386 &	0.6797 & 14.73          & -                            & -                   
  & 0.7143 &	0.7177 &	0.5216 &	\underline{18.23} & - & - \\
 & Korhonen  & 0.8042	& 0.8092	& 0.6066 &   16.16    & 0.867                        & 0.186   
 &0.6748 &	0.7003 &	0.4898 &	15.92 & 0.865 &	0.113 \\
& UAP   & \underline{0.7221} & \underline{0.7171} & \underline{0.5203} & \underline{17.64} & 0.792    & 0.141
&0.7204	&0.7264 &	0.5322 &	18.12 & 0.650	& 0.159\\
& Zhang & 
0.8213 &	0.8048 &	0.6259	&	16.70  & 0.853   & 0.084  
& \underline{0.5614} &	\underline{0.5695} &	\underline{0.3983} &	16.71 & 0.791 &	0.113 \\ 
\multirow{-5}{*}{\small{CONTRIQUE}}& Ours  & \textbf{0.5705}	& \textbf{0.6063}	& \textbf{0.4046}	& \textbf{19.40} & {0.901}   & {0.040}    
&\textbf{0.0667} &	\textbf{0.1234} &	\textbf{0.0509} &	\textbf{21.26} & 0.896	& 0.078\\ \bottomrule[0.4mm]
\end{tabular}
}
\label{tab_all}
\end{table*}
\subsubsection{Compared Attack Methods} To compare with the existing method, we choose the only black-box attack method for NR-IQA from Korhonen and You~\cite{korhonen2022adversarial}. It utilizes a variant of ResNet50~\cite{he2016deep} as its substitute model. We use a learning rate of 2 to generate the adversarial examples with its public code from the authors and mark it as {Korhonen}. For a comprehensive comparison, two white-box attack methods trained with substitute models are compared as transfer-based black-box methods, marked as UAP~\cite{shumitskaya2022universal} and Zhang~\cite{zhang2022perceptual}. For UAP, we use the perturbation generated with the substitute model PaQ-2-PiQ~\cite{ying2020patches}. The amplitude for the perturbation is set to 0.024. For Zhang~\cite{zhang2022perceptual}, we re-generate adversarial examples with substitute model DBCNN~\cite{zhang2020blind} with the perceptual constraint of LPIPS~\cite{zhang2018unreasonable}, and Lagrangian multiplier $\lambda=9\times10^6$.

\subsection{Attacking Results}
We present the prediction performance of NR-IQA models before (marked as Original) and after the attack in Table~\ref{tab_all}.
Our method has superior attack effectiveness under the premise of maintaining good invisibility. It consistently leads to substantial performance degradation across not only correlation metrics but also the MAE. Specifically, the attack on CONTRIQUE within the CLIVE dataset results in an SROCC reduction from above $0.7$ to under $0.1$, indicating a substantial disruption in the order relationship within the image set.
Meanwhile, Zhang presents unstable attack performances with failure in attacking SFA on the LIVE dataset. For instance, the SROCC for SFA unexpectedly increases from under $0.85$ before the attack to above $0.85$ after the attack. The Korhonen method performs a better MAE value than our attack in targeting SFA because the substitute model it used is similar to the model used in SFA. But our method still achieves better SROCC/PLCC/KROCC performance. 

Attack performance compared with the $R$ metric is shown in Table~\ref{tab_r}. Our method shows superior results, either ranking the best or second-best in attacking all four NR-IQA models.  It achieves an $R$ value of $0.826$ attacking DBCNN on the CLIVE dataset, significantly surpassing the second-best method by over $0.14$. 
When considering the SFA, we observe that the Korhonen method attains a lower $R$ value on both datasets. This can be explained by the similarity between the substitute model employed by Korhonen and the model utilized in SFA.

\begin{table}[htbp]
\centering
\setlength\tabcolsep{3pt}
\caption{Attack performance comparison with the $R$ metric. The best and second performances are marked with \textbf{bold} and \underline{underline}. }
\resizebox{0.8\linewidth}{!}{
\begin{tabular}{ll>{\centering\arraybackslash}p{1.4cm}>{\centering\arraybackslash}p{1.4cm}}
\toprule[0.4mm]
\multirow{2}{*}{\begin{tabular}[c]{@{}l@{}}Attacked\\ NR-IQA\end{tabular}} & \multirow{2}{*}{\begin{tabular}[c]{@{}l@{}}Attack\\ Method\end{tabular}} & \multicolumn{2}{c}{Attack Performance ($R$$\downarrow$)}                 \\ \cline{3-4} 
& & LIVE           & CLIVE                \\ \midrule
\multirow{4}{*}{DBCNN} & Korhonen             & 1.571          & \underline{0.982}          \\
& UAP  & \underline{0.942}    & 1.805                \\
& Zhang & 1.544          & 1.185                \\
 & Ours  & \textbf{0.869} & \textbf{0.826}       \\ \hline
\multirow{4}{*}{HyperIQA} & Korhonen             & {0.987}          & \textbf{0.762}       \\
 & UAP   & \textbf{0.870}    & 1.153                \\
 & Zhang & 1.403          & 1.220                \\
 & Ours  & \underline{0.982} & \underline{0.890}          \\ \hline
\multirow{4}{*}{SFA}   & Korhonen             & \textbf{0.654} & \textbf{0.495}       \\
 & UAP   & 1.168          & {0.866}          \\
 & Zhang & 1.510          & 1.107                \\
 & Ours  & \underline{0.843}    & \underline{0.887}                \\ \hline
\multirow{4}{*}{CONTRIQUE} & Korhonen             & 1.516          & 1.123                \\
 & UAP   & \underline{1.256}          & \underline{1.038}               \\
 & Zhang & 1.537          & 1.064                \\
 & Ours & \textbf{1.058} & \textbf{0.851} \\
\bottomrule[0.4mm]
\end{tabular}}
\label{tab_r}
\end{table}

For the robustness of NR-IQA models, all models present the vulnerability to black-box attacks on both synthetic distortions in the LIVE dataset and authentic distortions in the CLIVE dataset, which alarms the necessity to explore the security of NR-IQA. Among the four NR-IQA models, DBCNN and SFA suffer the most performance degradation with low correlation metrics, MAE, and $R$ value. Meanwhile, NR-IQA models are less robust against attacks on images with authentic distortions compared to synthetic ones. This could be attributed to the more complex and variable patterns in authentic distortions, presenting an easier target for attacks. It is worthy to be further explored in future work.%

\subsection{Analysis of Adaptive Iterative Score Boundaries}\label{sec_boundary}
In this subsection, we explore the impact of proposed adaptive iterative score boundaries, namely ``{iterative boundaries}'' and ``{adaptive optimization}''. The experiments are conducted on attacking the DBCNN model within the LIVE dataset.

\subsubsection{Iterative Boundaries Analysis} Adversarial examples are generated with different numbers of adaptive score boundaries. In part B of Table~\ref{tab_boundary}, the number $N$ of score boundaries directly affects the intensity of the attack. Totally, an increase in $N$ correlates with a heightened attack intensity. This correlation underscores the direct impact of iterative numbers on the intensity of adversarial examples generated. %

\begin{table}[!thpb]%
\setlength\tabcolsep{3.5pt}
\centering
\caption{Black-Box attack performance with different settings of score boundaries. Experiments are conducted on attacking the DBCNN model within the LIVE dataset.}
\resizebox{0.9\linewidth}{!}{
\begin{tabular}{lccc|cc} %
\toprule[0.4mm]
& \multirow{2}{*}{Setting} &\multicolumn{2}{c|}{Attack Performance} & \multicolumn{2}{c}{Invisibility}\\ \cline{3-6}
&      & SROCC$\downarrow$   & MAE$\uparrow$  & SSIM$\uparrow$  & LPIPS$\downarrow$               \\ \hline %
A. Original & - & 0.9529 & 9.79 & - & - \\ \hline
\multirow{4}{*}{\begin{tabular}[c]{@{}l@{}}B. \#Score \\ Boundaries\end{tabular}}              & $N=5$                                       & 0.8147 &	16.55   & 0.900	& 0.055 \\ %
 & $N=10$ & 0.5042 &	 20.50   & 0.864 & 0.108 \\ %
& $N=20$ & 0.3985	 & {20.93} & 0.867 & 0.082 \\ %
& $N=40$ & \textbf{0.3963} & \textbf{21.74} & 0.855  & 0.090  \\ \hline %
\multirow{2}{*}{\begin{tabular}[c]{@{}l@{}} C. Adaptive \\ Boundaries\end{tabular}} 
& Fixed & 0.8849 &	15.17  & 0.919 & 0.061 \\ %
& Adaptive & \textbf{0.3985}	 & \textbf{20.93} & 0.867 & 0.082 \\ \bottomrule[0.4mm] %
\end{tabular}
}%
\label{tab_boundary}
\end{table}

\subsubsection{Adaptive Optimization Analysis} Fixed boundaries with $\gamma_i=0.01i, i=1,...,N$ are attempt comparing with adaptive optimization. $N$ is set to 20. Results are shown in part C of Table \ref{tab_boundary}. In contrasting fixed boundaries with adaptive optimization, we observe that fixed boundaries, despite resulting in less perceptible perturbations, generally yield inferior attack performance. Adaptive optimization, where $\gamma_i$ is dynamically adjusted across images and iterations, distinctly enhances attack intensity. This adaptability ensures that each attack is optimally tailored to a different target image, facilitating a stronger attack intensity.

\subsection{Analysis of Different Settings for Initial Perturbation $\boldsymbol{d}_\text{tex}$} \label{sec_dtex}
For the initial perturbation $\boldsymbol{d}_\text{tex}$ in Sec.~\ref{sec:single}, the high-frequency information $\boldsymbol{I}_\text{freq}$ is employed.
Different settings for $\boldsymbol{d}_\text{tex}$ are explored by attacking the DBCNN model within the LIVE dataset.

\subsubsection{Different Options for $\boldsymbol{d_\text{tex}}$}\label{sec_dtex_option} 
\begin{figure*}[htbp]%
  \centering
   \includegraphics[width=0.65\linewidth]{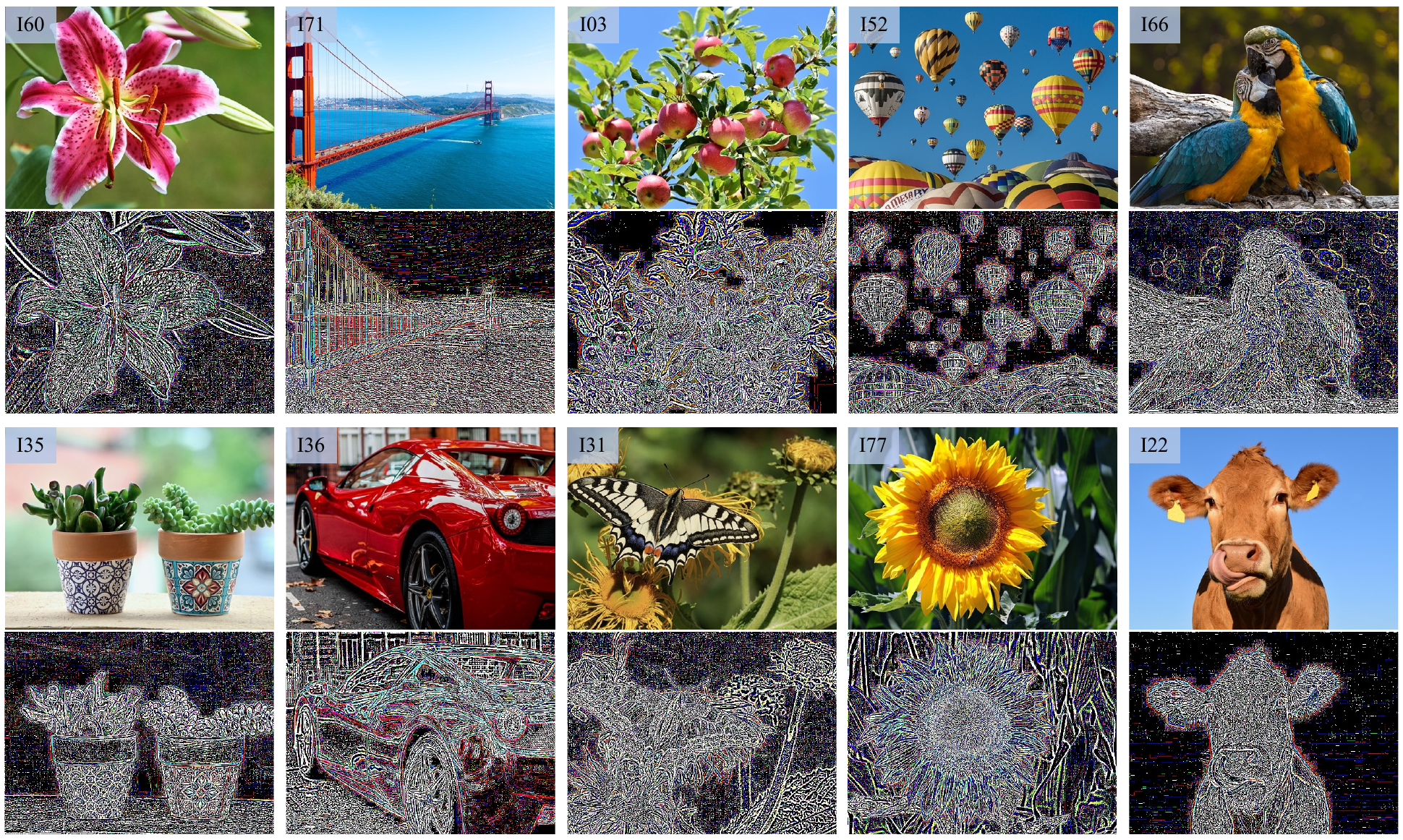}
   \caption{High-quality images randomly selected from the KADID-10k dataset in the first/third row, their names are labeled in the top left corner. The corresponding extracted texture and noised images are in the second/fourth row.}
   \label{fig:kadid}
\end{figure*}

There are different options for $\boldsymbol{d}_\text{tex}$, like natural images $\boldsymbol{I}_\text{nat}$ and high-frequency information $\boldsymbol{I}_\text{hfre}$. 
We verify the effect of different options for $\boldsymbol{d_\text{tex}}$: utilizing $\boldsymbol{I}_\text{nat}$ or $\boldsymbol{I}_\text{hfre}$ as $\boldsymbol{d_\text{tex}}$. 
For $\boldsymbol{I}_\text{nat}$, four high-quality images are randomly selected from the KADID-10k dataset, as shown in the first two images in the first row of Fig.~\ref{fig:kadid}, and one of them is randomly chosen as the $\boldsymbol{d}_\text{tex}$ in a single-step attack. For $\boldsymbol{I}_\text{hfre}$, the high-frequency information extracted from the first two images in Fig.~\ref{fig:kadid} are used, and one of them is randomly chosen as the $\boldsymbol{d}_\text{tex}$ in a single-step attack. The attack performance is shown in the first part of Table \ref{tab_dtex}. Utilizing $\boldsymbol{I}_\text{nat}$ as the $\boldsymbol{d}_\text{tex}$ proves to be effective to some degree, which makes SROCC decrease by around $0.15$. Using $\boldsymbol{I}_\text{hfre}$ extracted from the same natural images achieves much better attack performance. It is attributed to the role of image texture and sparse noise in high-frequency images. 

\begin{table}[tpb]%
\setlength\tabcolsep{4pt}
\centering
\caption{Black-Box attack performance with different settings for $\boldsymbol{d}_\text{tex}$. Experiments are conducted on attacking the DBCNN model within the LIVE dataset.}
\resizebox{\linewidth}{!}{
\begin{tabular}{lccc|cc} %
\toprule[0.4mm]
& \multirow{2}{*}{Setting} &\multicolumn{2}{c|}{Attack Performance} & \multicolumn{2}{c}{Invisibility}\\ \cline{3-6}
&      & SROCC$\downarrow$   & MAE$\uparrow$  & SSIM$\uparrow$  & LPIPS$\downarrow$               \\ \hline %
A. Original & - & 0.9529 & 9.79 & - & - \\ \hline
\multirow{2}{*}{\begin{tabular}[c]{@{}l@{}}B. Options\\ of $\boldsymbol{d}_\text{tex}$\end{tabular}}  
& \multicolumn{1}{c}{$\boldsymbol{I}_\text{nat}$} & 0.7925	&	16.95 & 0.907  & 0.039 \\ %
& $\boldsymbol{I}_\text{hfre}$%
& \textbf{0.3985}	 & \textbf{20.93} & 0.867 & 0.082 \\ \hline %
\multirow{3}{*}{\begin{tabular}[c]{@{}l@{}}C.\\Components\\of $\boldsymbol{d}_\text{tex}$\end{tabular}}  
& Sparse Noise  & 0.7930 & 15.62  & 0.928 & 0.030 \\ %
 & Image Texture  & 0.4086 & 20.67 & 0.869 & 0.102\\ %
& Both 
& \textbf{0.3985} & \textbf{20.93} & 0.867 & 0.082 \\ \bottomrule[0.4mm] %
\end{tabular}
}%
\label{tab_dtex}
\end{table}

\begin{table}[htb]
\centering
\caption{The mean and standard deviation of attack performance matrices and invisibility matrices with 10 $\boldsymbol{I}_\text{hfre}$ related to different contents. The experiments are conducted on attacking the DBCNN model within the LIVE dataset.} %
\begin{tabular}{ccc|cc}
\toprule[0.4mm]
 \multirow{2}{*}{\makecell{High-Frequency \\ Image Name}} &\multicolumn{2}{c|}{Attack Performance} & \multicolumn{2}{c}{Invisibility}\\ \cline{2-5}
 & SROCC$\downarrow$   & MAE$\uparrow$  & SSIM$\uparrow$  & LPIPS$\downarrow$               \\ \hline 
Original  & 0.9529 & 9.79 & -     & -      \\ \hline
I60   & 0.5848     & 19.89 & 0.890 & 0.070 \\
I71   & 0.5263     & 19.40 & 0.876 & 0.070 \\
I03    & 0.5456     & 20.04 & 0.885 & 0.068 \\
I52   & 0.6494     & 18.46 & 0.898 & 0.055 \\
I66   & 0.5339     & 20.18 & 0.887 & 0.069 \\
I35   & 0.6461     & 19.35 & 0.899 & 0.065 \\
I36   & 0.5964     & 19.44 & 0.883 & 0.068 \\
I31   & 0.5757     & 20.23 & 0.894 & 0.066 \\
I77   & 0.4843     & 20.61 & 0.877 & 0.070 \\
I22   & 0.4121     & 21.00 & 0.894 & 0.067 \\ \hline
Mean & 0.5555     & 19.86 & 0.888 & 0.067 \\
Std  & 0.0723 & 0.73  & 0.008 & 0.004  \\
\bottomrule[0.4mm]  
\end{tabular}
\label{tab_content}
\end{table}

\subsubsection{Different Contents Related to $\boldsymbol{I}_\text{hfre}$} 
When using $\boldsymbol{I}_\text{hfre}=g(\boldsymbol{I}_\text{nat})$ as $\boldsymbol{d}_\text{tex}$, whether the image content of $\boldsymbol{I}_\text{nat}$ influences attack performance and invisibility of adversarial attacks? %
We randomly select 10 high-quality images $\boldsymbol{I}_\text{nat}$ with different image contents from the KADID-10K dataset, as shown in Fig.~\ref{fig:kadid}, which vary from scenarios and contents. Their corresponding $\boldsymbol{I}_\text{hfre}$ are regarded as ten different $\boldsymbol{d}_\text{tex}$. The same test set within the LIVE dataset is attacked ten times with these different $\boldsymbol{d}_\text{tex}$ respectively. As shown in Table~\ref{tab_content}, the standard deviations of SROCC and MAE are merely under $0.1$ and $1$, which implies the variations among different $\boldsymbol{d}_\text{tex}$ are remarkably low. Meanwhile, the variation of SSIM and LPIPS are also minimal. It implies the image content of high-quality images has little effect on the performance and invisibility of the attack.%

\subsubsection{Image Texture vs. Sparse Noise in {$\boldsymbol{I}_\text{hfre}$}}
As shown in Fig.~\ref{fig:kadid}, $\boldsymbol{I}_\text{hfre}$ is composed of two components: image texture and sparse noise.
When using $\boldsymbol{I}_\text{hfre}$ as $\boldsymbol{d}_\text{tex}$, what are the effectiveness of different components of $\boldsymbol{I}_\text{hfre}$?
To examine it, we segment\footnote{The segmentation was done manually on https://segment-anything.com.} the high-freq image $\boldsymbol{I}_\text{hfre}$ into image texture and sparse noise, and regard them as the initial perturbation $\boldsymbol{d}_\text{tex}$ respectively. 
The high-quality images I60 and I71 are selected from the KADID-10k dataset~\cite{lin2019kadid} as $\boldsymbol{I}_\text{nat}$. 
The result of attacking DBCNN on the LIVE dataset is shown in Table~\ref{tab_dtex}. 
Both image texture and sparse noise contained in $\boldsymbol{I}_\text{hfre}$ are effective for attacking. Image texture witnesses a small SROCC and larger MAE value, which implies it plays a more important role. Meanwhile, when both image texture and sparse noise are utilized (i.e. the whole $\boldsymbol{I}_\text{hfre}$ is used), the attack performance achieves the best among them. It confirms the role of image texture and sparse noise in high-frequency images when used as the initial attack direction.

\subsection{Ablation Study}
To examine the effectiveness of different parts of our attack method, we conduct a detailed performance analysis by attacking the DBCNN model within the LIVE dataset for different settings in Table~\ref{tab_n}. The original performance on unattacked images is shown in part A of Table~\ref{tab_n}. 

\begin{table}[!thpb]%
\setlength\tabcolsep{4pt}
\centering
\caption{Black-Box attack performance with different settings. The experiments are conducted on attacking the DBCNN model within the LIVE dataset.}
\resizebox{\linewidth}{!}{
\begin{tabular}{lccc|cc} %
\toprule[0.4mm]
& \multirow{2}{*}{Setting} &\multicolumn{2}{c|}{Attack Performance} & \multicolumn{2}{c}{Invisibility}\\ \cline{3-6}
&      & SROCC$\downarrow$   & MAE$\uparrow$  & SSIM$\uparrow$  & LPIPS$\downarrow$               \\ \hline %
A. Original & - & 0.9529 & 9.79 & - & - \\ \hline
\multirow{2}{*}{\begin{tabular}[c]{@{}l@{}}B. Operation on\\ Initial Attack \\ Direction\end{tabular}} & Edge  & 0.4650 &	20.81  & 0.878 & 0.093 \\ %
 & Sal.   & 0.7207 &	17.14 & 0.900 & 0.076 \\ %
& Edge+Sal.%
& {0.3985}	& {20.93} & 0.867 & 0.082 \\ \hline %
\multirow{2}{*}{\begin{tabular}[c]{@{}l@{}}C. Strategy for\\Optimization\end{tabular}}  
& \multicolumn{1}{c}{Incr.} & 0.8801	&	20.54 & 0.893  & 0.098 \\ %
& GL
& {0.3985}	& {20.93} & 0.867 & 0.082 \\ \hline %
\multirow{2}{*}{\begin{tabular}[c]{@{}l@{}}D. Constraint \\ of JND\end{tabular}}     & w/o JND & -0.2989 &  29.91  &{0.729}    & {0.212} \\ %
 & with JND  & {0.3985}	& {20.93} & 0.867 & 0.082 \\ \bottomrule[0.4mm] %

\end{tabular}
}%
\label{tab_n}
\end{table}

\subsubsection{Effect of $\textbf{Mask}_{\text{edge}\cup\text{sal}}$ in Sec.~\ref{sec:single}}
The employment of edge mask and saliency mask operation confine perturbations to specific regions, as detailed in sec.~\ref{sec:single}. The effect of these operations (marked as Edge and Sal. respectively) are explored in part B of Table~\ref{tab_n}.
Compared to the original performance, both operations lead to an effective attack. Only using the edge mask performs a more important role with an MAE above $20$. Utilizing both masks together yields the most potent attack. %

\subsubsection{Effect of GL Optimization in Sec.~\ref{sec:GL}} The GL optimization strategy, formulated in Eq.~\eqref{eq2}, aims to attack high-quality images to obtain lower quality scores and vice versa. To verify the effect of GL optimization, we compare it with an increasing strategy, as recommended in \cite{korhonen2022adversarial,shumitskaya2022universal}. The increasing strategy aims to obtain higher quality scores for all attacked images (marked as Incr.). The results are shown in part C of Table~\ref{tab_n}. Both Incr. and GL strategies provide effective attacks compared to the original performance, with a larger MAE value compared to the original performance. Meanwhile, the employment of GL optimization provides a dramatic decline in SROCC. It is primarily attributed to, in GL optimization, the different strategies for higher/lower quality images significantly altering the ranking of predicted scores after the attack.
\begin{figure*}[!t]
    \centering
    \includegraphics[width=0.8\linewidth]{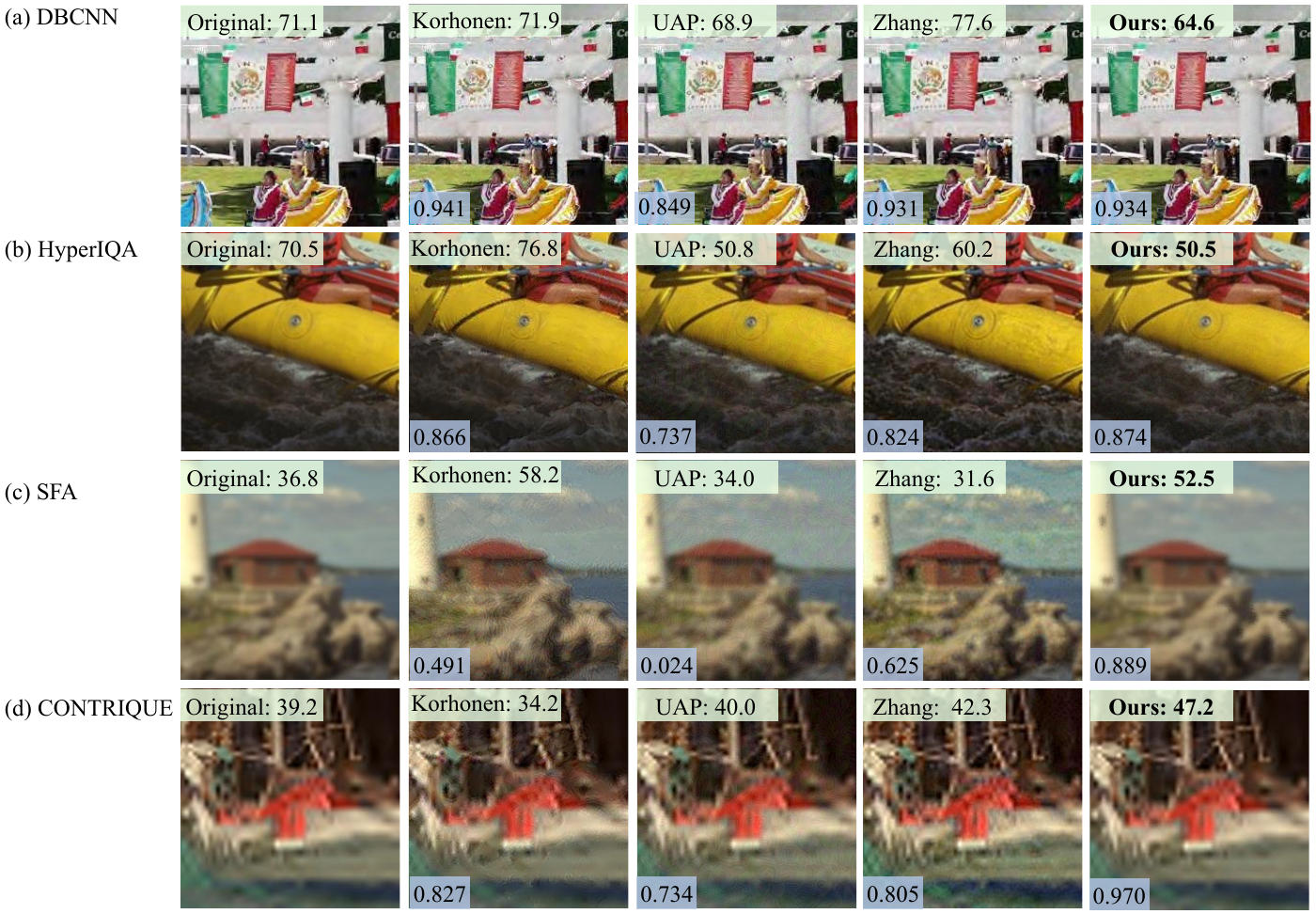}
    \caption{Adversarial examples from different attack methods on the LIVE dataset. Each row shows one attacked NR-IQA model, and each column corresponds to one attack method. The predicted quality score is shown on the top of each image. The SSIM value between the adversarial example and the original image is shown at the bottom of each image.
    Our method is marked with \textbf{bold}.}
    \label{fig:vis_live}
\end{figure*}

\begin{figure*}[!t]
    \centering
    \includegraphics[width=0.8\linewidth]{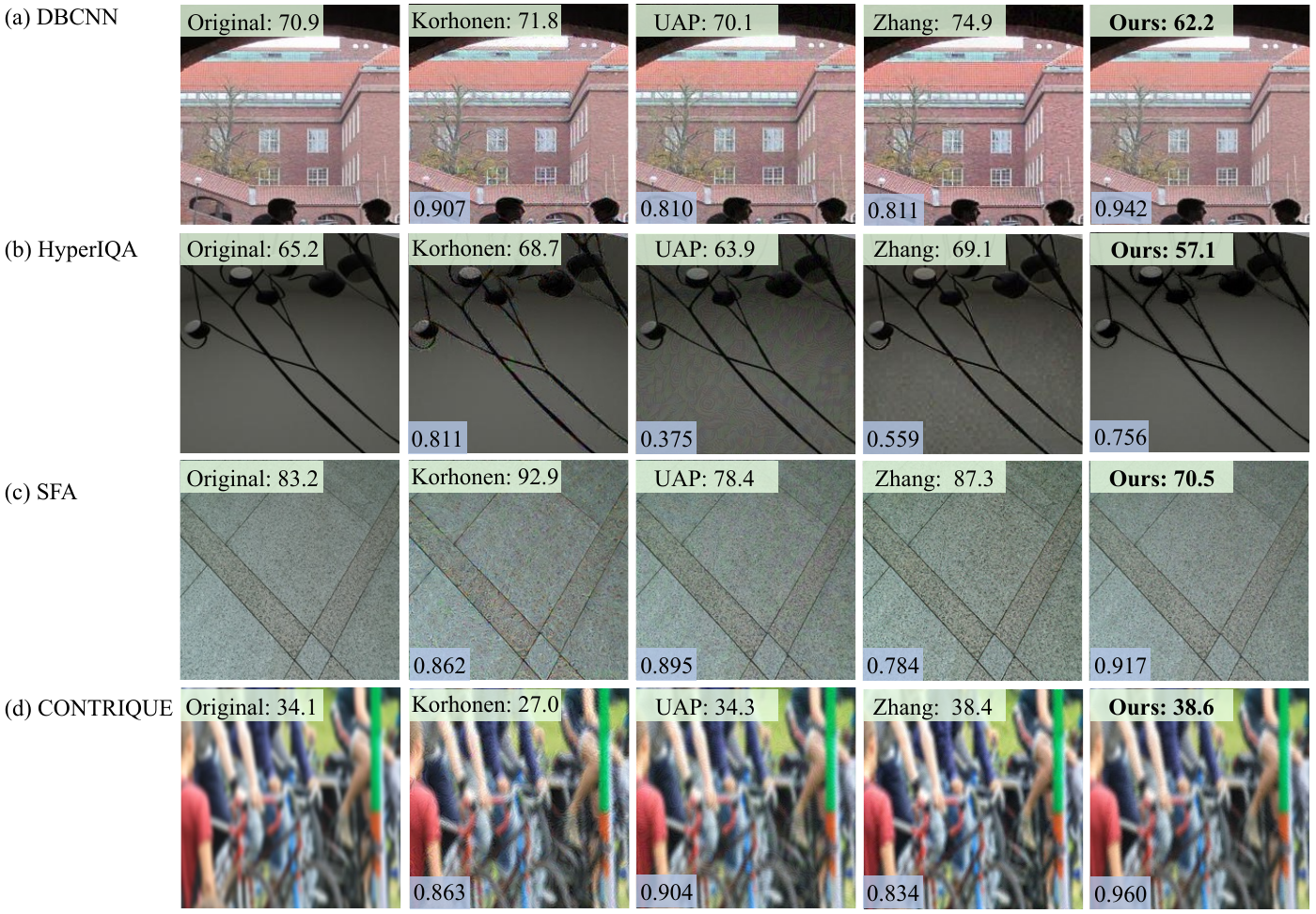}
    \caption{Adversarial examples from different attack methods on the CLIVE dataset. Each row shows one attacked NR-IQA model, and each column corresponds to one attack method. The predicted quality score is shown on the top of each image. The SSIM value between the adversarial example and the original image is shown at the bottom of each image. Our method is marked with \textbf{bold}.}
    \label{fig:vis_clive}
\end{figure*}

\subsubsection{Effect of JND Constraint in Sec.~\ref{sec:single}} The JND constraint in Sec.~\ref{sec:single} is designed to preserve the quality of adversarial examples, ensuring their perceptual invisibility. As shown in part D of Table \ref{tab_n}, the absence of this constraint significantly compromises the attack's invisibility, highlighting the JND constraint's critical role in balancing effectiveness with imperceptibility.

\subsection{Visualization of Adversarial Examples}
The imperceptibility of perturbation is an important part of the adversarial attack. We guarantee it with the JND constraint. Meanwhile, to verify the effectiveness of the JND constraint, we calculated the perceptual similarity between the adversarial example and its original image with two metrics. 

For an intuitive exhibition, we show the visualization of adversarial examples generated on the LIVE and CLIVE datasets in Fig.~\ref{fig:vis_live} and Fig.~\ref{fig:vis_clive}, respectively. 
It is noticeable our adversarial examples show good similarity with the original images. The perturbations generated by our method are more concentrated in the high-frequency region, like the rocks in Fig.~\ref{fig:vis_live}~(c), and the black wires in Fig.~\ref{fig:vis_clive}~(b). 
Other methods tend to have more perturbations in the low-frequency region, like the sky in Fig.~\ref{fig:vis_live}~(c), and the wall in Fig.~\ref{fig:vis_clive}~(b), which are easier to be captured by the human eye. 
It implies the necessity of our constraint with $\textbf{Mask}_{\text{edge}\cup\text{sal}}$.  On the other hand, our attack performs a good invisibility on the blurred images like Fig.~\ref{fig:vis_live}~(c) and Fig.~\ref{fig:vis_clive}~(d). In contrast, the other three attack methods generate textures that appear unnatural in these images, resulting in poor invisibility.

\subsection{Comparison with Other Attack Methods}
\subsubsection{Comparison with Black-Box Attack in Image Classification Task}
Although there are many existing query-based black-box attack methods proposed for the image classification task. They face the problem that classification boundaries can not be transferred to NR-IQA tasks and the querying without any prior is inefficient. 

To compare our attack method with existing black-box attacks in classification tasks, we adopt a classical query-based black-box attack method SimBA~\cite{guo2019simple} to the attack of NR-IQA. SimBA uses pixel-wise search to decrease the probability of correct classification predicted by the attacked model. To adopt SimBA for attacking NR-IQA, we utilize the optimization objective in Eq.~\eqref{eq2}. The loss function is:
\begin{equation}
\begin{gathered}
    J(\boldsymbol{x}_0,\boldsymbol{d})=S(\boldsymbol{x}_0)*(f(\boldsymbol{x}_0+\boldsymbol{d})- f\left(\boldsymbol{x}_0\right)), \\
    \text { s.t. } \boldsymbol{x}_0+\boldsymbol{d}\in \mathcal{D}_{\text{JND}}(\mathbf{x_0}),
\label{eq_simba_loss}
\end{gathered}
\end{equation}
where 
\begin{equation}
    \begin{aligned}
    S(\boldsymbol{x})=\left\{
    \begin{array}{ll}
    -1, &\boldsymbol{x} \in \mathcal{I}_l \\
    1, &\boldsymbol{x} \in \mathcal{I}_h
    \end{array}.\right.
    \end{aligned}
    \label{eq_s}
\end{equation}
The adopted algorithm SimBA-IQA is shown in Algorthm~\ref{alg:algorithm_simba}. We use step size $\epsilon=20/255$ for SimBA-IQA. The Cartesian basis of orthogonal search vectors $Q$ is selected, which corresponds to each iteration we are increasing or decreasing one color of a single randomly chosen pixel. 

\begin{figure}[htbp]
\centering
\includegraphics[width=.85 \linewidth]{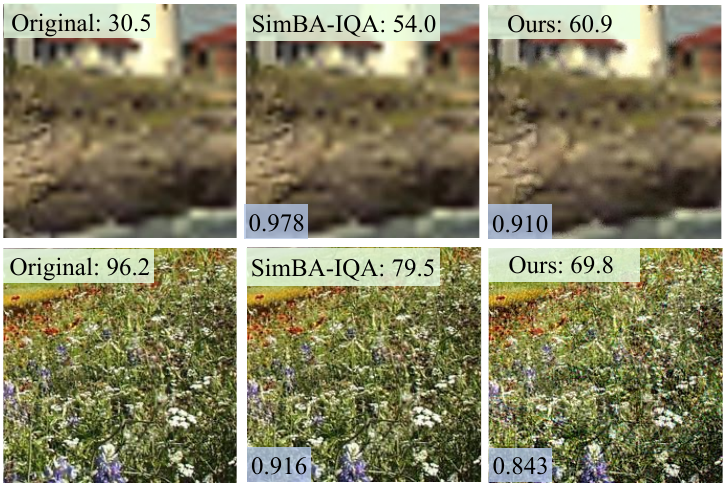}
\caption{Adversarial examples generated from SimBA-IQA and our attack against DBCNN on the LIVE dataset. The quality score predicted by the DBCNN/SSIM value is on the top/bottom of the images. Although the SSIM value of SimBA-IQA is a little higher, we can hardly distinguish the difference between adversarial samples generated by ours and SimBA-IQA.}
\label{vis_simba}
\end{figure}

The performance when attacking DBCNN on the LIVE dataset is shown in Table~\ref{tab_simba}. Within the same number of queries, SimBA-IQA obtains substantially weaker attack intensity than our attack. With increasing the number of queries to $20,000$, the attack intensity of SimBA-IQA is improved, but it is still weaker than our attack, which is only $8,000$ query times used. Though SimBA-IQA guarantees a better SSIM value than ours, visual results (Fig.~\ref{vis_simba}) indicate that human observers can not notice the difference between ours and SimBA-IQA.
SimBA-IQA performs a low efficiency in attacking NR-IQA, cause it does not consider the prior information for attacking NR-IQA, and it uses the pixel-wise attack mechanism. Meanwhile, the set of initial attack directions and the optimization with SurFree in our attack method guarantee both effective attack intensity and a low number of queries. 

\begin{algorithm}[tb]
\caption{Algorithm for SimBA-IQA}
\label{alg:algorithm_simba}
\textbf{Input}: Original image $\boldsymbol{x}_{0}$, the set of orthogonal search vectors $Q$, and the step size $\epsilon$\\
\textbf{Output}: Adversarial example $\boldsymbol{x}$
\begin{algorithmic}[1] %
\STATE $\boldsymbol{\boldsymbol{d}}\leftarrow0$
\STATE $\hat{J}\leftarrow J(\boldsymbol{x}_0,\boldsymbol{0})$
\WHILE{$T<T_\text{max}$}
\STATE {Pick randomly without replacement: $\boldsymbol{q}\in Q$}
\FOR{$\alpha\in\{\epsilon,-\epsilon\}$}
\STATE $J'=J(\boldsymbol{x}_0,\alpha\boldsymbol{q})$
\IF{$J'<\hat{J}$}
\STATE $\boldsymbol{d}\leftarrow\boldsymbol{d}+\alpha\boldsymbol{q}$
\STATE $\hat{J}\leftarrow J'$
\ENDIF
\ENDFOR
\ENDWHILE
\RETURN $\boldsymbol{x}_0+\boldsymbol{d}$
\end{algorithmic}
\end{algorithm}

\begin{table}[htpb]
\centering
\caption{Performance comparison with black-box method SimBA-IQA. Experiments are conducted on attacking the DBCNN model within the LIVE dataset.}
\resizebox{\linewidth}{!}{
\begin{tabular}{lccc|cc}
\toprule
   \multirow{2}{*}{\makecell{Attack \\ Method}} & \multirow{2}{*}{\makecell{\#Queries}} & \multicolumn{2}{c}{Attack Performance} & \multicolumn{2}{c}{Invisibility} \\ \cline{3-6} 
  &   & SROCC$\downarrow$ & MAE$\uparrow$ & SSIM$\uparrow$  & LPIPS$\downarrow$   \\ \hline
  Original & - & 0.9157 & 12.07 & - & - \\ \hline
SimBA-IQA & 8,000 & {0.7713} & 16.23   &  0.956 & 0.006      \\
SimBA-IQA & 20,000 & 0.7127 & 17.77   &  0.946 &  0.007 \\
Ours & 8,000  &  \textbf{0.3985}	& \textbf{20.93} & 0.867 & 0.082 \\
\bottomrule
\end{tabular}}
\label{tab_simba}
\end{table}

\subsubsection{Comparison with Black-Box Attack in VQA Task}
In the field of VQA, Zhang~\etal \cite{zhang2023vulnerabilities} propose a black-box attack method for videos. It employs a patch-based random search method, to assign a universal perturbation $\textbf{m}$ to randomly selected patches across video frames. Each element in $\textbf{m}$ is independently sampled from a discrete set $\{+\gamma,-\gamma\}$.
Additionally, a score-reversed boundary loss is used to mislead the VQA model to inaccurately predict quality scores: lower scores for high-quality videos and vice versa. The difference between \cite{zhang2023vulnerabilities} and our attack is the perturbation generation strategy. Unlike a universal perturbation, our strategy leverages specific characteristics of the image content, such as texture or saliency information, to craft perturbations, which are more efficient and effective.

To evaluate our proposed attack against this method, we adopt it with the NR-IQA task, denoted as PatchAttack-IQA. In this adaptation, the height and width of the universal perturbation $\textbf{m}$ are set to $14\times14$, with $\gamma$ set to $12/255$. 
The performance when attacking DBCNN on the LIVE dataset is shown in Table~\ref{tab_patch}. PatchAttack-IQA shows effective attack performance with a decline of SROCC with near $0.08$ when the number of queries is $8,000$. Meanwhile, under the same number of queries, our attack shows a better attack performance with a decline of SROCC exceeding $0.6$. With increasing the number of queries to $40,000$, the attack
intensity of PatchAttack-IQA increased but was still weaker than our method. Though PatchAttack-IQA utilizes a score-reversed boundary loss similar to our GL optimization, its performance in the NR-IQA context is notably limited. This limitation can be attributed to its simplistic perturbation generation strategy, where no prior information on images is taken into consideration. Conversely, our method shows a more efficient attack with better invisibility.

\begin{table}[htpb]
\centering
\caption{Performance comparison with the black-box method PatchAttack-IQA. Experiments are conducted on attacking the DBCNN model within the LIVE dataset.}
\resizebox{\linewidth}{!}{
\begin{tabular}{lccc|cc}
\toprule
   \multirow{2}{*}{\makecell{Attack \\ Method}} & \multirow{2}{*}{\makecell{\#Queries}} & \multicolumn{2}{c}{Attack Performance} & \multicolumn{2}{c}{Invisibility} \\ \cline{3-6} 
  &   & SROCC$\downarrow$ & MAE$\uparrow$ & SSIM$\uparrow$  & LPIPS$\downarrow$   \\ \hline
  Original & - & 0.9529 & 9.79 & - & - \\ \hline
PatchAttack-IQA & $8,000$ & {0.8739} & 12.06   &  0.786 & 0.136      \\
PatchAttack-IQA & $40,000$ & {0.8579}  & 11.18   &  0.784 & 0.140      \\
Ours & $8,000$ &  \textbf{0.3985}	& \textbf{20.93} & 0.867 & 0.082 \\
\bottomrule
\end{tabular}}
\label{tab_patch}
\end{table}

\section{Conclusion}
In this paper, we propose the query-based black-box attack for NR-IQA for the first time. We propose the definition of score boundary and leverage an adaptive iterative approach with multiple score boundaries. Meanwhile, the
design of attack directions ensures the effectiveness and invisibility of the attack. With the attack, the robustness of four NR-IQA models is examined. It reveals the vulnerability of NR-IQA models to black-box attacks and gives a clue for the exploration of the robustness of NR-IQA models.

\bibliographystyle{IEEEtran}
\bibliography{arxiv}

\vfill

\end{document}